\documentclass{article}
\usepackage{ijcai21}

\usepackage{algorithm}
\usepackage{algorithmic}
\usepackage{amsmath,amssymb,amsfonts, amsthm}
\usepackage{array}
\usepackage{booktabs}
\usepackage[small]{caption}
\usepackage{diagbox}
\usepackage{graphicx}
\usepackage[hidelinks]{hyperref}
\usepackage[utf8]{inputenc}
\usepackage{multirow}
\usepackage{mathtools}
\usepackage{subcaption}
\usepackage{soul}
\usepackage{times}
\usepackage{url}
\usepackage{xcolor}

\newcolumntype{P}[1]{>{\centering\arraybackslash}m{#1}}

\title{Land Cover Mapping in Limited Labels Scenario: A Survey}

\author{
Rahul Ghosh$^1$\and
Xiaowei Jia$^2$\and
Vipin Kumar$^1$
\affiliations
$^1$University of Minnesota\\
$^2$University of Pittsburgh
\emails
\{ghosh128, kumar001\}@umn.edu, xiaowei@pitt.edu
}

\begin{document}

\maketitle

\begin{abstract}
Land cover mapping is essential for monitoring global environmental change and managing natural resources. Unfortunately, traditional classification models are plagued by limited training data available in existing land cover products and data heterogeneity over space and time. In this survey, we  provide  a  structured  and comprehensive overview of challenges in land cover mapping and machine learning methods used to address these problems. We also discuss the gaps and opportunities that exist for advancing research in this promising direction.  

\end{abstract}

\section{Introduction}
\label{sec:Introduction}
Global demand for land resources to support human livelihoods and well-being through food, fiber, energy and living space will continue to grow in response to the population expansion and socioeconomic development. This poses a great challenge to the human society, given the increasing competition for land from the need to maintain other essential ecosystem services. Addressing this challenge will require timely information on land use and land cover (LULC) changes, e.g., the conversion of forest to farmland or plantations, the loss of productive cropland due to urbanization, and the degradation of soil due to inappropriate management practices. 

Advances in Earth observation technologies have led to the acquisition of vast amounts of accurate, timely, and reliable remote sensing (RS) data that can be used for monitoring changes on the Earth system at a global scale. These rich datasets provide tremendous potential to study the nature and impact of global changes due to natural processes and human activities that shape the physical landscape and environmental quality of our planet.

Supervised classification methods, especially recent deep learning approaches, have achieved significant success in commercial applications in Natural Language Processing (NLP) and Computer Vision (CV) domain, where large training data is available. Given the promise provided by these approaches, there is a growing interest for using these techniques in automated land cover mapping at large scale through the analysis of RS data~\cite{zhu2017deep}. However, off-the-shelf supervised classification methods have limited utility due to challenges that are unique to environmental applications. Supervised machine learning algorithms, e.g., advanced deep neural networks, require sufficient labeled training instances which are representative of the test data. Such training data is often scarce in land cover applications given high manual labor and material cost required in manual labeling (e.g., visual inspection) and field study. 
This is further exacerbated by the high-dimensional nature of spatio-temporal remote sensing data.  Moreover,  land covers commonly show much variability across space and time, e.g., the same crop can look different in different years and in different regions due to variability in weather conditions and farming practice. Additionally, the availability of multiple RS data sources acquired at different spatial and temporal resolutions, and other heterogeneous data, e.g., elevation, thermal anomalies, and night-time light intensity, provides unique algorithmic challenges that need to be addressed. 
The goal of this survey is to summarize recent developments in ML methods that are being used to address these challenges in the context of LULC. We  focus on scenarios with a broad range of limited labels, e.g., scarce labels or imperfect labels, which also makes it challenging to model the heterogeneity, the large number of classes, and the high-dimensional spatio-temporal data. 
Additionally, we discuss the gaps and opportunities that exist for advancing research in this promising direction. 

We organize the paper as follows. Section~\ref{sec:Objective} describes different challenges that can be achieved using latest ML techniques for learning with limited data. Then, Section~\ref{sec:Methods} discusses ML methods that researchers are using to address these challenges. Table~\ref{tab:Taxonomy}  categorizes the work surveyed in this paper by challenge and approach. We hope that this survey will also be valuable for domain scientists who are interested in exploring the use of ML to enhance modeling in mapping target land covers. Although the focus of this survey is largely on land cover application, the ideas carry over to a wide range of spatio-temporal data.

\section{Challenges}
\label{sec:Objective}

In this section, we provide an overview of a diverse set of challenges that have been encountered in land cover mapping.  

\subsection{Heterogeneity in space and time}
\label{sec:heterogeneity}
One major challenge in land cover mapping is the presence of a rich variety in the land cover characteristics in different regions and at different times, i.e., data heterogeneity. For example, crops can be planted under different soil types, precipitation, and other weather conditions. Farmers also use different land management practices (e.g., conservation tillage, low phosphorous), future crop rotation plan, and local social exchanges. All these factors can result in the heterogeneity over different places and different years.

Traditional ML methods that learn a single model of target land covers over the entire population of training data fail to account for this heterogeneity of the data in space and time, thus resulting in poor prediction performance when applied across diverse regions and times. Additionally, supervised learning algorithms require sufficient availability of labeled instances for training, which are expected to be adequately representative of the test data. However, training labels are often only available for specific regions and years due to the high associated costs of obtaining gold-standard labels. This results in poor prediction performance when the learned model is generalized to a larger region or different years.

\subsection{Large number of classes}
\label{sec:unseen_class}
Land cover products commonly involve a large number of land cover classes. For example, USDA Crop Data Layer~\cite{cdl} has defined over 150 land cover classes in the area of the US. Accurately mapping all these land covers requires the access to training data of all the classes. However, some land cover classes can be very rare. Such scenario is very common in the fine-grained land cover mapping, e.g., crop classification, where we may not have labeled instances of many such rare crops. Moreover, training labels are often only available for specific regions and years. Hence, unseen/novel land cover classes may appear when the trained model is applied to other regions or other years. Traditional ML models are designed to make classification over a fixed set of classes and cannot identify new classes that are not included in the training set. 

\subsection{Paucity of labeled data}
Most land cover labels are created by domain experts through visual inspection or field study. Given the substantial human labor and material cost required in the labeling process, it is often difficult to obtain large training data for land cover mapping. This poses a serious challenge for effectively training advanced machine learning models. Recent deep learning models, that have shown high performance in a variety of commercial applications, usually involve a large number of parameters and thus require large labeled training data.

This issue can be further exacerbated by the severe skewness of land covers. In particular, there can be much fewer samples for certain minority land cover classes compared with dominant classes such as forest, grassland and barren land. Although some minority classes have very few amount of training samples, detecting these classes can be critical for understanding the impact of human practice and land cover changes to the environment. For example, only small number of farmers have planted cover crops in the winter time so it is challenging to train a cover crop classifier using such limited data. However, planting cover crops is highly encouraged by the government since they can hep maintain the soil nutrient and prevent water contamination. Hence, the capacity to monitor the extent of cover crops is urgently needed. 

\subsection{Imperfect labels}
Training labels provided in many existing land cover products are noisy due to a number of reasons. First, inconsistent standard used in the annotation process. In a prior study for plantations,~\cite{petersen2016mapping} divided Landsat images into a grid of 20 × 20 km and visually scanned each gridded structure for multiple types of plantations with the assistance of forest gain and loss information.  Such an annotation process results in less than 80\% precision according to their independent assessment.  Moreover,  when applied to map larger regions, this approach may require multiple observers to delineate tree crop plantations, and observers are likely to be inconsistent with one another. 
Second, many labeling information is present at a much coarser scale, e.g., in the form of bounding boxes (approximate location), image-level labels (whether the input image contains objects), annotated polygons over the region of interest. Furthermore, many land cover products only provide labels for objects of interest, and the negative samples are left unmarked. Lastly, although some proxy labels are available, e.g., active fire for wild fire mapping and night-time light intensity for urban area mapping, they often either over-estimate or under-estimate target land covers. 

\subsection{High-dimensional spatio-temporal data}
Many RS datasets contain reflectance values from multiple spectral bands and thus are high dimensional. Given the power of deep learning models in automatically extracting hidden features, researchers have been using these models in learning new data representations from RS data. Research along this direction is of great importance since it may provide new opportunities to understand the relation between spectral features and land covers. In particular, deep learning models have the promise to extract new vegetation indices from multi-spectral reflectance values. Such indices can be potentially more robust and representative compared with existing vegetation indices, such as Normalized Difference Vegetation Index (NDVI), which are defined based on domain knowledge.  Moreover, the learned representation, together with the latest interpretation methods, can provide useful insights about the change of spectral data in response to different land covers. 

Besides interpretability, the obtained representations can also be used for a variety of downstream applications. For example, these representations can be used to measure the similarity amongst different land covers. Also, one can gather representations learned from different regions and use their distance to better quantify spatial heterogeneity.

\section{ML Methods}
\label{sec:Methods}

\subsection{Zero-shot Learning}
Zero-shot learning (ZSL) methods have been proposed for classification in a special scenario, where we consider two sets of classes -- \textit{seen classes} ($\mathcal{S}$), for which labeled instances are present, and \textit{unseen classes} ($\mathcal{U}$), for which there are no labeled instances. These two set of classes are disjoint and the aim is to learn a classifier that can classify the instances in the testing set into classes belonging to \textit{unseen classes}. To capture the relationship between seen and unseen classes, ZSL methods create a real-valued semantic space where each class $Y \in {\mathcal{S} \cup \mathcal{U}}$ (both seen and unseen) has a corresponding vector representation, which is also referred to as the class prototype. To create this semantic space, some form of auxiliary information is often necessary to describe the relation amongst classes in ${\mathcal{S} \cup \mathcal{U}}$. The class prototype along with the training data are then used to learn the zero-shot classifier. ZSL methods have been used in several land cover studies given their ability to automatically classify rare or unseen classes. Existing ZSL methods can be categorized based on the type of semantic spaces and the methods used to learn the zero-shot classifier~\cite{wang2019survey}.

Recent literature has demonstrated the capacity of ZSL in levering additional text data to extract semantic space for unseen land cover classes~\cite{li2017zero,gui2018generalized,sumbul2017fine,pradhan2020unseen}. For example,~\cite{li2017zero} introduced the idea of zero-shot scene classification using high spectral resolution (HSR) images. They first constructed a semantic space for embedding class prototype using the word2vec model~\cite{mikolov2013distributed} trained on Wikipedia. Then they further created a semantic directed graph over both seen and unseen classes and assign the edge weight as the distance between each pair classes based on their class prototypes. Once establishing the relationship between the seen and unseen, they fine-tune a GoogLeNet model~\cite{szegedy2015going} on the training data. The combination of GoogLeNet and the semantic graph enables producing the probability of a new instance belonging to each of seen or unseen classes.  In another work,~\cite{sumbul2017fine} extended ZSL methods for fine-grained tree classification by using a combination of label embeddings learned from text data via word2vec, visual embeddings of trees, and hierarchy embeddings based on scientific classification of tree species. Such combination allows creating a semantic space that leverage the benefits of both domain knowledge and learning based methods.

An extension of this idea in the context of land-cover mapping is presented in~\cite{jia2017incremental}. Instead of using auxiliary text data, authors in this work create the semantic space using sequences of land cover changes collected from other regions in the world. They have shown the effectiveness of this method in identifying deforestation and forest fires without using extra text data. Additionally,~\cite{chen2019generalized} used the human-defined attributes as semantic information for knowledge transfer between seen classes and unseen classes and used a projection method to learn the zero-shot classifier. Specifically, they  projected image features and label attributes to the same space and train a zero-shot classifier, which can  then classify the unseen classes by finding the closest class-attribute in the projected space.

Although notable developments of ZSL models have been published for various CV tasks, their applications to RS are still at a nascent stage with the focus on small regions and small sets of target land covers. Moreover, most of existing methods create label-semantic space using the word2vec model, which was originally designed for text embedding and thus are not able to fully explore land cover characteristics. Hence, there is a need for new mechanism to create a semantic space that captures the properties of the classes and also includes the domain knowledge of the hierarchy of these classes. One potential method is the combination-based method of ZSL, where the idea is to construct the classifier for unseen classes by the combination of classifiers for basic elements that are used to constitute the classes. For example, in crop classification, one can create individual models to learn the basic elements during a crop cycle like sowing of seeds, growing season, harvesting period, texture and combining them for building the semantic space guided by the domain knowledge.

\begin{figure}
    \centering
    \includegraphics[width=\linewidth]{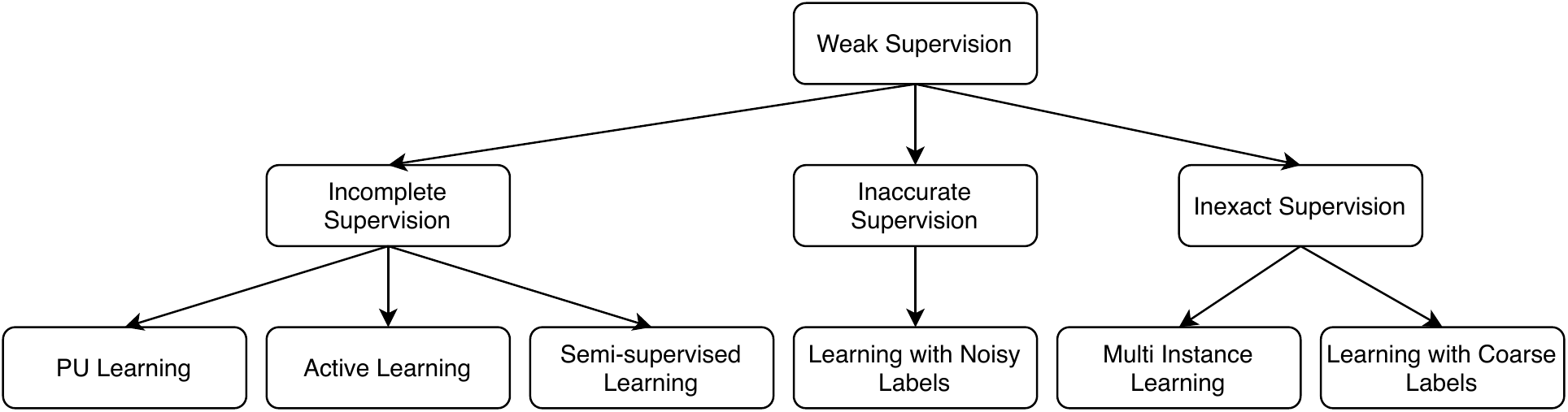}
    \vspace{-.15in}
    \caption{Weak supervision taxonomy.}
    \label{fig:Weak supervision taxonomy}
    \vspace{-.15in}
\end{figure}

\subsection{Weakly supervised Learning}
Weakly supervised learning (WSL) is an area of machine learning covering a variety of studies that attempt to construct predictive models by learning with different forms of weak supervision. In particular, there are  three types of weak supervision~\cite{zhou2018brief}: incomplete, inexact and inaccurate. An illustration of these three types along with the learning methodologies used in each of them is shown in Fig.~\ref{fig:Weak supervision taxonomy}. Several research works have demonstrated the benefits of such techniques on RS data for handling the challenges posed by weak supervision. 

ML methods that handle incomplete supervision, e.g., only target land covers are labeled, falls under the Positive-Unlabeled (PU) Learning family~\cite{bekker2020learning}. ~\cite{wei2020scribble} proposed a scribble based deep learning method for road surface extraction. Specifically, using the scribble based PU road labels from crowd sources such as GPS or Open Street Maps (OSM)~\cite{haklay2008openstreetmap}, negative samples (non-road pixels) are identified using buffers around the scribbles. These non-road pixels along with the scribble based road labels are used to train a semantic segmentation model. Another scenario of incomplete supervision arises when along with a few labeled samples, there is exists large unlabeled data. Semi-supervised learning~\cite{van2020survey} methods are typically used to handle this scenario.~\cite{han2020lightweight} adopted the Mean Teacher model~\cite{tarvainen2017mean} which is a popular semi-supervised technique to train their backbone ResNet~\cite{he2016deep} model using a few manually annotated samples and unlabeled samples for economic scales prediction.

The availability of large-scale crowd sourced information offers opportunities for building ML models for land cover mapping tasks for which expert-annotated labels are not available. However, crowd sourced data are often plagued by noise and thus require the ability to deal with inaccurate supervision.~\cite{sheehan2018learning} showed results on classifying satellite images using weak labels extracted from Wikipedia. Similarly,~\cite{sun2019leveraging} used crowd sourced GPS data for road extraction from aerial imagery. Apart from crowd sourced labels, other satellite products such as night-time lights also form a noisy proxy for several tasks such as urbanization and economic development. In their study,~\cite{han2020learning} showed the usability of night-time lights as a data rich proxy for identifying economic development from satellite imagery.

For pixel-wise land cover mapping tasks, coarse-level labels or image-level labels are often much easier to collect than pixel-level labels. Several methods have been proposed to leverage coarse-level labels for pixel-wise land cover mapping~\cite{zhang2019hierarchical,schmitt2020weakly,wang2020weakly}. In particular,~\cite{schmitt2020weakly} discussed the challenges and opportunities of using coarse-level and image-level weak labels for land cover mapping. Many of these approaches are inspired by image segmentation approaches using image-level labels~\cite{zhang2016weakly,fu2018wsf}.

Although most of the existing works have focused on single source of weak supervision, combining independent sources of weak supervision in a multi-task framework can force the models to learn from each other. For instance, for mapping urban areas the different sources of weak supervision can be night-time lights (coarse labels) and crowd sourced OSM labels (noisy labels). 

\subsection{Self-supervised Learning}
One common approach for dealing with limited availability of labeled datasets is to pre-train an ML model on existing large labeled datasets for a related problem, and then refine it using a small number of labeled samples for the problem of interest. For example, models for image recognition are first trained using large-scale datasets like ImageNet~\cite{deng2009imagenet} and then are fine-tuned on the limited-size dataset for the downstream task~\cite{huh2016makes}. However, such approaches cannot be used for RS due to two key reasons; a) Spectral bands available in RS satellite data are much richer than RGB data in ImageNet (e.g., Sentinel data has 12 bands); b) large scale labeled data for land covers are not available. 

Self-supervised learning (SSL)~\cite{jing2020self} is an alternative approach that learns feature representation from unlabeled images. Specifically, SSL methods aim to pre-train ML models using ``pre-text'' learning tasks related to target classes before fine-tuning ML models using small amount of observations. The central idea is to propose various pretext tasks for the network to solve, in the hope that the network will learn important feature representations related to specific problems, such as image classification, segmentation, and object detection. Some popular self-supervised methods in CV include inpainting patches~\cite{pathak2016context}, image colorization~\cite{zhang2016colorful}, and solving image-jigsaw~\cite{noroozi2016unsupervised}. 

Recently, researchers have started using SSL for land cover mapping. One popular approach is to pre-train the ML model by optimizing the reconstruction performance on unlabeled data. For example,~\cite{kemker2018low} introduced the reconstruction loss for each hidden layer of a stacked convolutional autoencoder for unsupervised training of the model. Specifically, they used a symmetric encoder-decoder framework where the feed-forward encoder segments the satellite images and the decoder network reconstructs the original input back from the compressed representations of the encoder. This method has shown improved predictive performance on RS datasets that have small annotated training data.

Many existing works have also adapted pre-text tasks that are widely used in CV to RS applications and have shown success in improving the performance. For example,~\cite{singh2018self} adopted the inpainting pretext task and showed its superiority over other autoencoder-based self-supervised methods in learning representation for segmenting land covers in several applications. In addition to the inpainting task, the authors propose to use an adversarial training scheme, where an adversarial network, called the coach network, is tasked to increase the difficulty of the inpainting task gradually so that the semantic inpainting model do not overfit to a single type of corruption. In~\cite{tao2020remote}, the authors compared image inpainting, image-jigsaw and contrastive learning~\cite{chen2020simple} based pretext tasks for the RS land cover classification and present the results on three datasets. Similarly,~\cite{vincenzi2020color} use an auto-encoder to predict the RGB channels given the other channels of the data as input. During the target classification tasks, the model has two independent branches which takes in the spectral bands and the RGB channels, respectively.

Another line of thought for pretext task, which is commonly used in the NLP domain, is representation learning based on context-similarity. The central idea is that words 
in similar contexts should have similar representations~\cite{mikolov2013distributed}. By redefining context as spatial neighborhoods, Tile2Vec~\cite{jean2019tile2vec} used this idea in RS domain where they promote the tiles that are close together to have similar representations than the tiles that are far apart.

SSL can also be used to extract shared information across multiple data sources. For example,~\cite{swope2019representation} proposed a method that creates different views of a given area and uses the InfoNCE loss to enforce consistency between the two views. To create two views of a multispectral image, standard data-augmentation techniques such as channel dropout, random flips, rotation, cropping and jittering the brightness and contrast are applied randomly. The consistency is enforced between the representations from different levels of a ResNet model. The authors show that the representations learned in this manner outperform the fully supervised ImageNet weights. Moreover, this multiview framework allows the fusion of multiple sources of data for joint representation learning using multi-modal data sources.

While much of the current research has focused on adopting ideas that have been successful in CV and NLP, there is also huge promise for designing new pretext learning tasks specific to land cover characteristics. One promising direction is to use domain knowledge, e.g., crop phenology, to define pre-training objectives. For example, neural networks can be pre-trained to predict several phenology-related values, including the slope of the NDVI curve, the maximum NDVI value, etc.   

\begin{table*}[]
    \tiny
    \centering
    \caption{Table of literature classified by challenges(rows) and method(columns).}
    \vspace{-.1in}
    \begin{tabular}{|P{1.4cm}||P{2cm}|P{2.5cm}|P{2cm}|P{2.5cm}|P{2cm}|P{2.5cm}|}
        \hline
          & ZSL & Weakly-supervised Learning & SSL & TL & Meta Learning & Feature Extraction \\
        \hline
        \hline
        Heterogeneity in space and time &  ~\cite{jia2017incremental} & ~\cite{tong2020land} 
        &  & \cite{jia2019classifying}\cite{tong2020land}\cite{islam2019semi} 
        &   &
        \\
        \hline
        Large number of classes & ~\cite{chen2019generalized} ~\cite{pradhan2020unseen} ~\cite{li2017zero} ~\cite{gui2018generalized} ~\cite{sumbul2017fine} &  &  & ~\cite{deng2020few} & & ~\cite{deng2020few} 
        \\
        \hline
        Paucity of labeled data  &  &  & ~\cite{xie2018unsupervised} ~\cite{tao2020remote} ~\cite{vincenzi2020color} ~\cite{singh2018self} 
        ~\cite{kemker2018low} ~\cite{romero2015unsupervised}& \cite{gadiraju2020comparative}\cite{marmanis2015deep}
        \cite{scott2017training}\cite{rostami2019deep}\cite{chen2018end} \cite{ayush2020generating}\cite{deng2020few} & \cite{zhang2021few}\cite{russwurm2020meta}\cite{li2020rs} 
        & \cite{xie2018unsupervised}\cite{gadiraju2020comparative}\cite{ayush2020generating} \cite{deng2020few}\cite{lu2017jm}\cite{romero2015unsupervised}
        \\
        \hline
        Imperfect labels  &  & \cite{sun2019leveraging}\cite{wang2020weakly}
        \cite{wei2020scribble}\cite{qiao2020simple}\cite{zhang2016weakly} \cite{zhang2019hierarchical}\cite{schmitt2020weakly}\cite{fu2018wsf} 
        \cite{han2020learning} &  & 
        \cite{xie2016transfer}\cite{najjar2017combining} &  & 
        \\
        \hline
        High-dimensional spatio-temporal data & \cite{demertzis2020geoai} & \cite{han2020lightweight}\cite{sheehan2018learning} & \cite{xie2018unsupervised}\cite{swope2019representation}\cite{jean2019tile2vec} & \cite{hu2015transferring}\cite{gargees2019deep}\cite{jia2019recurrent} & \cite{demertzis2020geoai} & \cite{xie2018unsupervised}\cite{gargees2019deep}\cite{demertzis2020geoai} \cite{albert2017using}\cite{liu2019siamese}\cite{yang2020tensor} 
        \\
        \hline
    \end{tabular}
    \label{tab:Taxonomy}
    \vspace{-.1in}
\end{table*}

\subsection{Transfer Learning}
Transfer learning (TL) is a class of methods which aim to improve a ML model on a target domain by transferring information from a related source domain. When there is limited labeled training data for the target domain, the existence of large scale datasets in a separate but related source domain can also provides the opportunity to transfer the knowledge gained from the source domain to the target task. TL has been extensively used in ML domain where models trained on large-scale datasets such as ImageNet have been successfully used as a starting point for solving the target task. 

Several studies have also shown the benefit of transferring the knowledge from large-scale CV datasets to RS data. In particular, many of these methods 
aim to deal with the paucity of labeled data~\cite{marmanis2015deep,hu2015transferring,najjar2017combining,chen2018end,scott2017training}. For example, \cite{hu2015transferring} present a comprehensive comparison of several network models 
pre-trained on large-scale datasets like ImageNet and Places~\cite{zhou2014learning}, and then fine-tuned on multiple RS benchmark datasets. Several other studies have shown the benefit of pre-training on ImageNet datasets in target tasks like road safety mapping~\cite{najjar2017combining}, airplane detection~\cite{chen2018end} and other land cover mapping datasets~\cite{scott2017training}. 

Another approach to TL is to pretrain the ML model using data rich proxy labels. This approach has several benefits: 1. Proxy labels can be very specific to the target class labels of interest (e.g., night time lights can be a good proxy for infrastructure available or level of poverty); 2. Transfer can be done meaningfully as they are more closely related to the target task than ImageNet. For instance, \cite{xie2016transfer} pre-train a ResNet model first using ImageNet dataset followed by night-time lights. The authors show that this method can achieve good performance when fine-tuned on limited labels of poverty prediction. As another example,~\cite{ayush2020generating} used the publicly available xView~\cite{lam2018xview} dataset, which is one of the largest and most diverse publicly available overhead imagery datasets for object detection. They also showed the success of transferring the models learned using xView as the source dataset to the target task of poverty prediction.

Several studies have also explored domain adaptation techniques for knowledge transfer. Domain adaptation aims to extract feature representation that is invariant across source and target domains and thus can potentially alleviate data heterogeneity.~\cite{rostami2019deep} used a domain adaptation method to transfer the knowledge from Electro-optical(EO) domain, which  has large number of labels available, to the synthetic aperture radar (SAR) image domain, where labelling can be difficult. They used two encoder networks to extract a shared invariant cross-domain embedding space such that the distributions of encoded EO and SAR data is minimized in this latent space.~\cite{jia2019classifying} utilized a cyclic generative adversarial network (GAN) model to perform the domain adaptation between source and target domains which enables them to address the heterogeneity in croplands and grasslands over different regions and different years. In similar lines,~\cite{islam2019semi} proposed a learning based domain adaptation technique using an adversarial loss to align marginal data distribution between a source region and a target region. Then a few labelled samples from the target domain is used to align class specific data distributions between the two domains, based on the contrastive semantic alignment loss.

Additionally, TL has been used to extract meaningful feature representations which can then be used for unsupervised grouping of land cover scenes.~\cite{gargees2019deep} used ImageNet and RSI-CB256~\cite{li2017rsi} benchmark dataset to train a land cover classification model. This model is then used as a feature extractor and the extracted features from RS images can be  used for clustering.


\subsection{Meta Learning}
Meta learning is a strategy that automatically learns how to adapt models quickly to different tasks. In meta learning,  a machine learning model gains experience from a distribution of related tasks and uses this experience to improve its future learning performance. Meta learning is performed in two levels: a) inner level where the model tries to solve a single task, and b) outer level where the meta model is updated using the experience gained by solving different tasks in the inner level. This enables the model to quickly learn a new task. 

Several works have shown the benefit of using meta learning methods for handling the paucity of labeled data in land cover mapping~\cite{zhang2021few,russwurm2020meta}.~\cite{russwurm2020meta} used model agnostic meta learning (MAML) for the problem of inductive transfer-learning, where the generalization is induced by a few labeled examples in the target domain. They show results on the Sen12MS dataset~\cite{schmitt2019sen12ms} and the DeepGlobe challenge.~\cite{li2020rs} proposed a novel meta learning framework for few-shot RS scene classification. Specifically, they generated different tasks sampled from a task family and learn a metric space that is tuned to perform well on new tasks.~\cite{zhang2021few} proposed a three stage meta learning based strategy for few-shot classification of aerial scene images. First, a feature extractor is trained using the base set and a standard cross-entropy loss. Second, the meta learning classifier is trained over a set of episodes in the meta-training stage while comparing the query features with the mean of support features. Finally, the meta learning classifier is tested on a set of novel set referred to as the meta-test stage.~\cite{demertzis2020geoai} proposed Model-Agnostic Meta-Ensemble Zero-shot Learning (MAME-ZSL) which exploits the MAML++ algorithm for the detection of unseen classes in land cover mapping.

Majority of the research works above have been on the MAML  direction of meta learning. The use of other meta learning algorithms such as metric-based meta learning is still rare, where the central idea is to automatically learn a similarity measure between two instances. If the similarity between the instances of target and source tasks can be successfully captured, it should lead in effective solving of the target tasks given its similarity to a particular source task. For instance, considering the identification of cashew plantations, corn crops and palm oil plantations as the (target, source$_1$ and source$_2$) task, the target task shares more similarity with source$_2$(both being plantations) than source$_1$. Thus, while identifying cashew plantations, the source task of identifying of palm oil plantation will be more beneficial. Another popular form of metric-based meta learning is prototypical networks~\cite{snell2017prototypical}, where a prototype feature vector is defined for every class using instance embeddings. Here, there is an opportunity to incorporate domain knowledge about the classes to create such prototypical vectors. 

\subsection{Feature extraction methods}
RS products often contain reflectance values from multiple spectral bands, e.g. MODIS has 7 spectral bands, Sentinel 2A has 13 spectral bands, and some hyper-spectral products have even more. The reflectance value on each spectral band can be sensitive to different atmosphere and ground conditions, e.g., clouds, leaf. Environmental scientists have previously created multiple indices, NDVI, Urban Index (UI), Normalized Difference Built-up Index (NDBI), based on their domain knowledge. These indices are essentially combinations of multiple spectral bands and have shown effectiveness in identifying specific land covers. However, the performance can be significantly degraded for distinguishing multiple similar land covers, e.g., multiple crops can have similar NDVI patterns if they are planted and harvested on close dates. 

The advent of new data-driven feature extraction methods have provided opportunities to extract more robust features from multiple spectral bands. Moreover, the state-of-the-art deep learning structures, such as convolutional neural network and recurrent neural network, also enable incorporating spatial and temporal context into the data representation. The extracted feature representation using these advanced techniques have shown improved performance in several applications. For example, prior work~\cite{deng2020few} has used a meta feature extractor based on Convolutional Neural Networks (CNNs) that extracts representative feature representations and enables learning with much fewer amounts of training samples. Moreover, such feature extraction can be also conducted using unlabeled data~\cite{romero2015unsupervised}.

Land cover changes are commonly driven by natural processes or human interventions, which have specific temporal and spatial patterns, e.g., deforestation continues to happen in nearby areas. This requires new feature extraction methods that incorporate both spatial and temporal relationships that are consistent to the change of weather conditions, human activities and their interactions. Additionally, feature extraction methods that can interpret such data dependencies will be highly valuable for better understanding land cover changes.

\section{Discussions, Future Directions and Concluding Remarks}

In this survey, we reviewed key challenges faced in land cover mapping and machine learning methods for addressing them. Given the advances in machine learning and deep learning, we anticipate these techniques to play an important role in future works on monitoring land cover changes. In Table~\ref{tab:Taxonomy}, we summarized relevant literature on different challenges using different methods. Also note that many boxes in  the  Table~\ref{tab:Taxonomy}  have no and very few entries,  many  of  which  represent  opportunities  for  future  work. For example, SSL has been widely used for pre-training ML models so that they need much fewer training samples for fine-tuning. There is also a great promise for using SSL for extracting new representations that can be used for downstream applications, such as clustering spatial regions.

Another promising direction is to combine RS with other data sources such as street-level images to jointly study land covers. This would require new multi-view and representation learning methods to handle data from different views and in different spatial and temporal resolutions~\cite{jia2019recurrent}.
Additionally, many environmental processes, e.g., the growth of crops in response to fertilizer, soil and water conditions, cannot be directly observed through RS data. Hence, it is important to combine RS data with underlying physical/bio-chemical processes. For example, the emerging research on physics-guided machine learning~\cite{willard2020integrating} has provided huge potential on this direction. 

\section{Acknowledgement}
This work was funded by the NSF awards 1838159 and 1739191. Rahul Ghosh is supported by UMII MNDrive Graduate Fellowship.

\bibliographystyle{named}
\bibliography{main}

\end{document}